\documentclass[10pt,journal,compsoc]{IEEEtran}
\usepackage{amsmath,amsfonts}
\usepackage{algorithmic}
\usepackage{algorithm}
\usepackage{array}
\usepackage[caption=false,font=normalsize,labelfont=sf,textfont=sf]{subfig}
\usepackage{textcomp}
\usepackage{stfloats}
\usepackage{url}
\usepackage{verbatim}
\usepackage{graphicx}
\usepackage{cite}
\usepackage{booktabs}
\usepackage{multirow}
\usepackage{xspace}
\newcommand{\ie}{{\it i.e.}\xspace}

\newcommand{\SeqFusion}{{\sc SeqFusion}\xspace}
\newcommand{\mphi}{{\boldsymbol{\phi}}}
\newcommand{\mmu}{{\boldsymbol{\mu}}}
\newcommand{\mtheta}{{\boldsymbol{\theta}}}
\newcommand{\sD}{\mathcal{D}}
\DeclareMathOperator*{\argmaxB}{argmax}

\begin{document}

\title{{\scshape{SeqFusion}}: Sequential Fusion of Pre-Trained Models for Zero-Shot Time-Series Forecasting}

\author{Ting-Ji Huang, Xu-Yang Chen, Han-Jia Ye
% \IEEEmembership{Fellow, IEEE}, Masaki Owari
\thanks{Ting-Ji Huang, Xu-Yang Chen, and Han-Jia Ye are with the State Key Laboratory for Novel Software Technology, Nanjing University, Nanjing, Jiangsu
210023, China. E-mail: {huangtj, chenxy, yehj}@lamda.nju.edu.cn.}
}


\markboth{Journal of \LaTeX\ Class Files,~Vol.~14, No.~8, August~2021}%
{Shell \MakeLowercase{\textit{et al.}}: A Sample Article Using IEEEtran.cls for IEEE Journals}


\maketitle

\begin{abstract}
Unlike traditional time-series forecasting methods that require extensive in-task data for training, \textit{zero-shot} forecasting can directly predict future values given a target time series without additional training data. Current zero-shot approaches primarily rely on pre-trained generalized models, with their performance often depending on the variety and relevance of the pre-training data, which can raise privacy concerns. Instead of collecting diverse pre-training data, we introduce \SeqFusion in this work, a novel framework that \textit{collects and fuses} \textit{diverse} \textit{pre-trained models (PTMs)} \textit{sequentially} for zero-shot forecasting. Based on the specific temporal characteristics of the target time series, \SeqFusion selects the most suitable PTMs from a batch of pre-collected PTMs, performs sequential predictions, and fuses all the predictions while using minimal data to protect privacy. Each of these PTMs specializes in different temporal patterns and forecasting tasks, allowing \SeqFusion to select by measuring distances in a shared representation space of the target time series with each PTM. Experiments demonstrate that \SeqFusion achieves competitive accuracy in zero-shot forecasting compared to state-of-the-art methods. Code is available at \url{https://github.com/Tingji2419/SeqFusion} .\looseness=-1

\end{abstract}

\begin{IEEEkeywords}
time series forecasting, zero-shot, pre-trained model.
\end{IEEEkeywords}

\section{Introduction}
\label{Section1}
Time-series forecasting is essential in various fields such as healthcare~\cite{penfold2013use}, finance~\cite{omer2020financial}, and environmental science~\cite{zahra2020transductive}, where it supports decision-making by predicting future trends based on historical data. Current predominant approaches, including traditional statistical methods~\cite{Hoptroff93, box1970distribution} and modern deep learning techniques~\cite{Zhou2021informer,iTransformer, patchtst,lu24the,onefitsall,softs,zijian24trans,gabriel22pay}, often require extensive in-task training data to capture complex temporal patterns, which are tailored to a particular field or environment similar to the target time series~\cite{kukjin2021deep}. However, in many real-world scenarios, there is often little or even no available training data~\cite{canton2019sales,vitor2019machine,simon2020}, which limits the effectiveness of these traditional and deep learning methods, reducing their practicality in real-world applications.\looseness=-1

\begin{figure}[htbp]
\centering
\includegraphics[width=0.95\columnwidth]{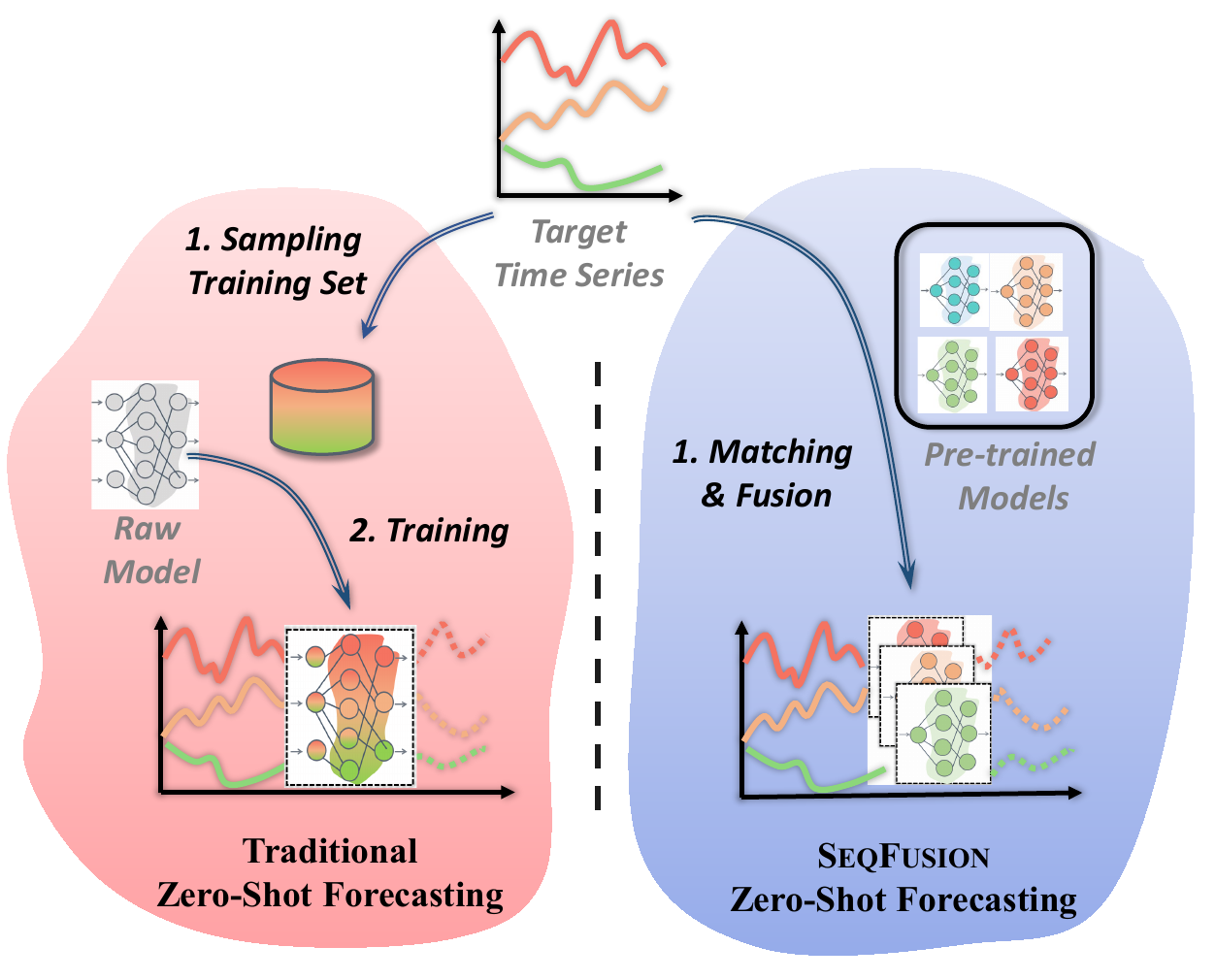}
  \caption{ Comparison of traditional zero-shot forecasting and \SeqFusion. Traditional methods rely on sampling data and training, while \SeqFusion leverages pre-trained models (PTMs) through matching the target time series to suitable PTMs and fusing their predictions.}
\label{fig:setting}
\vspace{-10pt}
\end{figure}

To address these challenges, \textit{zero-shot} forecasting has emerged as a promising solution~\cite{metanbeats}. This approach enables reliable predictions of target time series without in-task training data. Current studies mainly employ generalized models pre-trained on diverse datasets~\cite{TimeGPT, onefitsall,chronos}, which require the centralized collection of large and diverse data. However, these generalized models typically demand substantial memory storage and raise significant data privacy concerns~\cite{LLMprivacy, LiuDSRFL21}. Given that sharing pre-trained models (PTMs) or small demo datasets instead of entire source data is a common strategy for protecting data privacy~\cite{federated}, we pioneer an approach that utilizes a set of lightweight PTMs to help for zero-shot forecasting. \looseness=-1

%\textit{model zoo}
%说一下模型库的必要性
In this work, we propose \SeqFusion, a novel method that leverages pre-trained models (PTMs) for zero-shot time-series forecasting. Instead of collecting diverse data, we create a \textit{model zoo} containing a wide range of PTMs, mimicking the diverse set of PTMs that would naturally emerge from various time-series forecasting tasks in real-world applications~\cite{huggingface}. Each PTM in our model zoo is pre-trained on distinct datasets from various tasks, enabling them to specialize in different temporal patterns. 
Model zoo is common and acceptable in real-world applications~\cite{yang22a,modelspider,beimingwu}, such as models for stocks prediction~\cite{man23clustering}, where a company may have trained a large number of deep models for different time periods, different regions, or different domains. When faced with a new forecasting task, we would like to utilize these deep-learning PTMs if there is a lack of datasets. \looseness=-1

Collecting PTMs is beneficial for data privacy protection, as it reduces the risk of exposing sensitive information in the time series. Since PTMs are often black-box models, they do not reveal the underlying data used for their training. Thus, our approach provides the necessity and feasibility of sharing time-series PTMs in the future, which has already been a great success in the field of vision and natural language process~\cite{LLMprivacy,learnware}. Furthermore, storing PTMs reduces storage costs compared to storing raw time series data. PTMs require less storage space, and they can be easily shared and reused across different tasks and applications. 
% Furthermore, the storage requirements for PTMs are substantially lower than those for extensive raw time-series data. \looseness=-1

\SeqFusion employs a diverse collection of PTMs and a dynamic selection mechanism to identify the most suitable PTMs based on the specific characteristics of each variate in the target time series. With the selected PTMs, \SeqFusion forecasts sequentially and fuses the outputs of each PTM for every variate to obtain the final predictions. Each of these PTMs is pre-trained on different and diverse data, thus specializing in different temporal patterns and forecasting tasks. For instance, a model pre-trained on the Hospital~\cite{hospital} dataset might effectively forecast certain variates in the Illness~\footnote{\url{https://gis.cdc.gov/grasp/fluview/fluportaldashboard.html}} dataset due to their shared medical data characteristics and similar recording frequencies (See Table~\ref{tab:zoo_size} for more details). This flexibility is particularly useful for single-variate scenarios, where PTMs can be used interchangeably if the history length of two forecasting tasks is consistent, enhancing the practicality of the forecasting process.

The success of \SeqFusion hinges on two key factors: obtaining accurate characteristics of the target time series and the PTMs and aligning these characteristics to match suitable models with the target series. This is achieved through a time-series feature extractor, trained via an unsupervised learning process, ensuring that time series from the same dataset have similar representations, while also aligning PTMs' representations to reflect their transferability across related datasets. This dual alignment process is vital for ensuring that the selected PTMs are highly relevant and capable of producing accurate predictions. Using only a few data that do not harm privacy, we extract these representations and match appropriate PTMs with the target time series based on representation distance. 

Once suitable models are selected, \SeqFusion employs a sequential forecasting mechanism that predicts values segment by segment for each variate, and allows for aggregated predictions of the most relevant few PTMs. This direct prediction method, without additional training data, ensures flexibility and adaptability to various time series. Figure~\ref{fig:setting} illustrates the contrast between traditional forecasting, which relies on sampling from an in-task training set, and our method, which efficiently utilizes a set of PTMs without such sampling. Our contributions are:\looseness=-1

\begin{itemize}
\item  We introduce \SeqFusion, a novel framework that leverages pre-trained models (PTMs) for effective zero-shot time-series forecasting, eliminating the need for large-scale pre-training data.
\item  \SeqFusion selects suitable PTMs based on the target time series characteristics, performs sequential predictions, and fuses the predictions of all the selected PTMs to produce a robust final forecast.
\item Comprehensive experiments across diverse datasets demonstrate that \SeqFusion achieves competitive accuracy to state-of-the-art methods, particularly in environments with limited data and storage space.
\end{itemize}

\section{Related Work}
\label{Section2}

\noindent{\bf Time-Series Forcasting}. Time-series forecasting plays a vital role in various domains such as finance~\cite{omer2020financial}, weather prediction~\cite{zahra2020transductive}, and industrial manufacturing~\cite{kukjin2021deep,ming24a,olly22multi}. Classic methods like ARIMA~\cite{Hoptroff93} have long been the cornerstone of time-series forecasting. In recent years, the advent of deep learning methods has become popular in this area. Many methods utilize deep sequence models in an attempt to capture temporal dependencies and long-range dependencies in the series~\cite{Wu2021Autoformer,fedformer,iTransformer,han2024the}. Some recent studies leverage the large language models for time-series analysis that are based on billions of parameters and tremendous training data from various domains~\cite{TimeGPT,onefitsall}. However, these deep learning methods often require large amounts of relevant training data to achieve great performance and are difficult to perform in some data-lacking applications~\cite{forecastpfn,GruverFQW23}.

\noindent{\bf Unsupervised Representation of Time Series}. Unsupervised representation learning of time series aims to extract generalized representations from complex raw time series data without human supervision. Previous works apply joint clustering and feature learning approaches and show remarkable performance~\cite{ZhanX0OL20}, while others adopt deep models such as auto-encoders or sequence-to-sequence models to generate time-series representations~\cite{MaZLC19,MalhotraTVAS17}. Recently, self-supervised learning has emerged as a new paradigm, which leverages pretext tasks to autonomously generate labels by utilizing intrinsic information derived from the unlabeled data~\cite{LeiYVWD19,TonekaboniEG21,YueWDYHTX22, simmtm}. These inspire us to extract features of target time series for similar sequence matching to help forecasting in data-limited prediction tasks.

\noindent{\bf Zero-Shot Forecasting}. Zero-shot forecasting addresses the challenge of predicting unseen or novel time-series sequences without historical data. This paradigm shift is crucial for scenarios where traditional forecasting methods are impractical due to the absence of relevant training data~\cite{GruverFQW23}. Early works like Meta-N-BEATS~\cite{metanbeats} employ a model based on N-BEATS~\cite{nbeats} with flexible basis functions to generalize across different time-series domains. ForecastPFN~\cite{forecastpfn} predicts future series by constructing synthetic data and training a model with this, bypassing the need for real-world training data. Meanwhile, some previous work utilized pre-trained language models to generate predictions based on contextual understanding of time-series data~\cite{moment,TimeGPT, onefitsall,GruverFQW23, TTMs}. These methods leverage the inherent ability of language models to understand and generate sequences, thereby enabling them to make informed predictions even in the absence of direct historical data~\cite{chronos, MOIRAI}. However, the good performance of these zero-shot forecasting methods typically requires large and diverse pre-training data. In this work, we explore collecting diverse PTMs instead of pre-training data for zero-shot forecasting.

\noindent{\bf Pre-Trained Models for Forecasting}. Leveraging pre-trained models (PTMs) for forecasting tasks has gained momentum, especially as it allows models to generalize across diverse tasks with minimal task-specific training. In time-series contexts, PTMs can transfer learned temporal patterns from one domain to another, improving forecasting accuracy when data is sparse or task-specific patterns are hard to capture. Several recent studies demonstrate the efficacy of PTMs in forecasting~\cite{onefitsall,forecastpfn,chronos}. TimeGPT~\cite{TimeGPT}, a large-scale, closed-source model which is pre-trained on diverse time-series datasets. It captures complex temporal dependencies, making it highly effective in zero-shot forecasting scenarios. Chronos~\cite{chronos} tokenizes time-series data through scaling and quantization and uses transformer-based architectures to process these sequences. Moirai~\cite{MOIRAI}, a masked encoder-based transformer pre-trained on a large-scale dataset, which includes over 27 billion observations across nine domains. Its extensive pre-training enables Moirai to handle with an arbitrary number of variates for multivariate time series, and perform competitively in zero-shot forecasting compared with fully supervised models. TimesFM~\cite{TimesFM} employs a patched-decoder attention mechanism pre-trained on a large time-series corpus to capture long-range dependencies. This structure allows TimesFM to recognize nuanced temporal patterns, making it suitable for tasks requiring detailed temporal analysis. These works underline the potential of PTMs to serve as a shared repository or model zoo, where models trained on various domains and tasks can be selectively utilized for new forecasting applications, reducing the reliance on large datasets and extensive retraining. In this work, we build on the model zoo concept to facilitate forecasting in data-limited settings, aiming to make time-series PTMs more accessible and effective across various forecasting domains.

\section{Preliminary for \scshape{SeqFusion}}
We first describe the multivariate time-series forecasting problem under the zero-shot situation. Then we discuss a solution with PTMs and the sequential forecasting method used in \SeqFusion.

\subsection{Multivariate Zero-Shot Forecasting}
In multivariate time-series forecasting under the zero-shot setting, we are given a historical time series $\mathbf{X}=\{\mathbf{x}_1, ..., \mathbf{x}_C\} \in \mathbb{R}^{T \times C}$ with $C$ variates of a limited length $T$. The objective is to predict the future values of each variate $\mathbf{Y}=\{\mathbf{y}_{1}, ..., \mathbf{y}_{C}\} \in \mathbb{R}^{H \times C}$ over a horizon $H$ through a model $\mphi:\mathbb{R}^{T \times C} \rightarrow \mathbb{R}^{H \times C}$, where the length $T$ is insufficient to train a deep model even for one step~\cite{metanbeats}.

\begin{algorithm}[H]
\caption{Sequential Forecasting Process}\label{al:1}
\begin{algorithmic}[1]
\STATE \textbf{Input:} Time series data $\mathbf{x}$, model $\phi$, input length $T$, forecast horizon $H$, iteration steps $n$
\STATE \textbf{Output:} Forecast $\hat{\mathbf{y}}$ over horizon $H$
\STATE Initialize $\mathbf{x}_0 = \mathbf{x}$
\STATE Compute initial output $\mathbf{y}^0 = \phi(\mathbf{x}^0)$
\FOR{$t = 1$ \textbf{to} $n$}
    \STATE Concatenate inputs and previous outputs: $[\mathbf{x}^0, \mathbf{y}^0, \ldots, \mathbf{y}^{t-1}]$
    \STATE Trim concatenated data to maintain length $T$: $\mathbf{x}^t = \text{Trim}([\mathbf{x}^0, \mathbf{y}^0, \ldots, \mathbf{y}^{t-1}])$
    \STATE Apply model $\phi$ to get new output: $\mathbf{y}^t = \phi(\mathbf{x}^t)$
\ENDFOR
\STATE Concatenate all outputs: $[\mathbf{y}^0, \mathbf{y}^1, \ldots, \mathbf{y}^{n}]$
\STATE Trim concatenated outputs to match forecast horizon $H$: $\hat{\mathbf{y}} = \text{Trim}([\mathbf{y}^0, \mathbf{y}^1, \ldots, \mathbf{y}^{n}])$
\end{algorithmic}
\end{algorithm}

\subsection{Sequential Forecasting with PTMs}

When dealing with a target time-series forecasting task with limited input length, direct training a deep model may not be feasible due to data limitations. Instead of collecting diverse data to pre-train a generalized model, a feasible approach is directly utilizing a PTM to make zero-shot forecasting. 

Given a target time-series $\mathbf{X}$ and a PTM $\mphi:\mathbb{R}^{T \times C}\rightarrow\mathbb{R}^{h \times C}$ pre-trained on another task, if the length of $\mathbf{X}$ is the same as the input history length $T$ of $\mphi$ and the number of variates is also the same, we can use $\mphi$ to directly predict the future value $\hat{\mathbf{Y}}=\mphi(\mathbf{X})$. However, considering that the prediction length of the PTM $\mphi$ is often fixed to $h$, which is common in recent deep learning methods~\cite{Wu2021Autoformer,Zhou2021informer,DLinear}, it may not be directly applicable in a zero-shot setting, especially when the required horizon $H$ is longer than the prefix prediction length $h$.

To address this, we introduce a sequential forecasting approach for $n$-step ahead recursive forecasting~\cite{DirMO, souhaib2012a}. Specifically, for each variate, the task of forecasting a horizon $H$ is decomposed into $n= \lceil H/h \rceil$ forecasting blocks. For each block, we make multivariate predictions using $\mphi$, where the output of the previous module is truncated to a suitable input length for the model and used as input for the next module. Ultimately, we synthesize the output of each module and truncate it to $H$. Each yields a predicted value of $\hat{\mathbf{y}}_c$ for each variate. The steps of this sequential prediction are detailed in Algorithm~\ref{al:1}.\looseness=-1

% \begin{figure*}[t]
%     \begin{subfigure}[b]{0.43\linewidth}
%         \centering
%         \includegraphics[width=\linewidth]{fig/method-a.pdf}
%          \caption{Matching Pre-Trained Models for Time Series.}
%          \label{fig:method_a}
%     \end{subfigure}
%     \begin{subfigure}[b]{0.56\linewidth}
%     \centering
%         \includegraphics[width=\linewidth]{fig/method-b.pdf}
%          \caption{Zero-Shot Forecasting with \SeqFusion.}
%          \label{fig:method_b}
%     \end{subfigure}
%   \caption{Overview of \SeqFusion. (a) \SeqFusion collects diverse PTMs and selects the most suitable PTMs based on the characteristics of the target time series. This selection is based on the representations obtained from a general extractor, to provide a vector for measuring the similarity between the PTMs and each variate in the target time series. (b) \SeqFusion employs sequential forecasting with the selected PTMs for each variate, and fuses all predictions over all variates to generate final forecasts. The sequential forecasting process includes normalization, trimming, and the optional aggregated prediction of the most suitable PTMs, followed by concatenation and de-normalization to produce the final forecasts.}
%   \label{fig:method}
% \end{figure*}

\begin{figure*}[t]
\centering
\hspace{-0.2in}
\subfloat[Matching PTMs for Time Series.]{\includegraphics[width=2.7in]{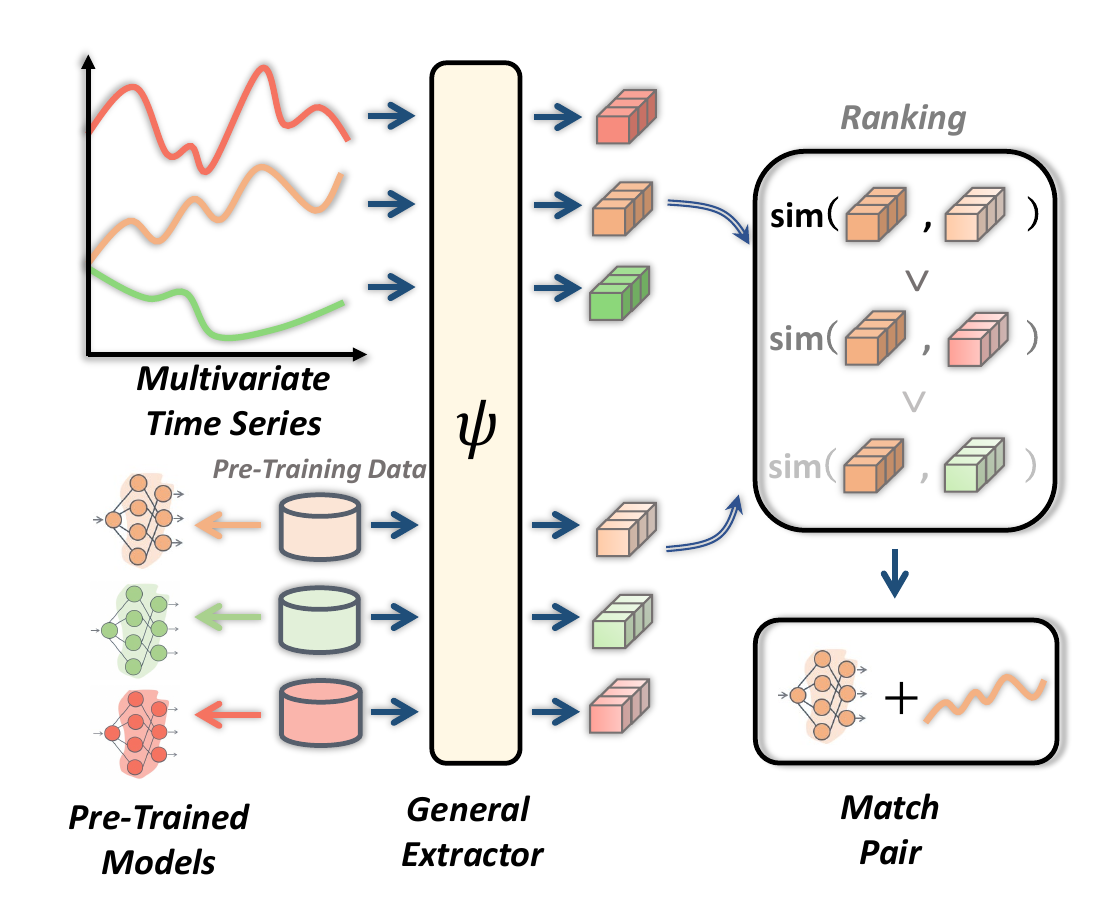}%
\label{fig:method_a}}
\hspace{0.2in}
\subfloat[Zero-Shot Forecasting with \SeqFusion.]{\includegraphics[width=3.7in]{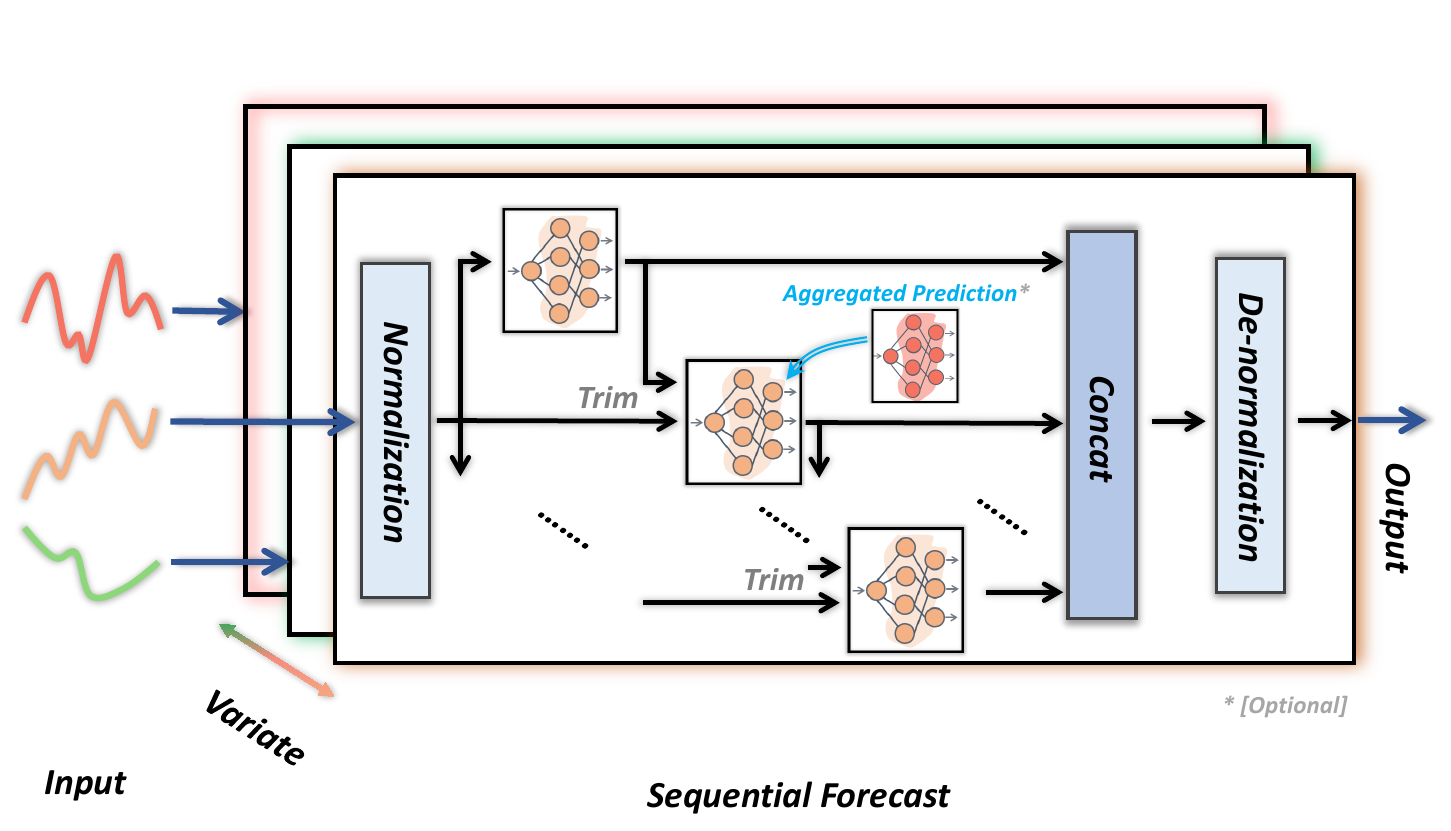}%
\label{fig:method_b}}
\hspace{-0.2in}
\caption{Overview of \SeqFusion. (a) \SeqFusion collects diverse PTMs and selects the most suitable PTMs based on the characteristics of the target time series. This selection is based on the representations obtained from a general extractor, to provide a vector for measuring the similarity between the PTMs and each variate in the target time series. (b) \SeqFusion employs sequential forecasting with the selected PTMs for each variate, and fuses all predictions over all variates to generate final forecasts. The sequential forecasting process includes normalization, trimming, and the optional aggregated prediction of the most suitable PTMs, followed by concatenation and de-normalization to produce the final forecasts.}
\label{fig:method}
\end{figure*}

\section{\scshape{SeqFusion}}
\label{Section4}

In this section, we first introduce the model zoo with diverse PTMs and the overall architecture of \SeqFusion for zero-shot time-series forecasting. We then describe the process of obtaining time-series representations, as well as incorporating time-series transferability to improve the semantics of the representations. Following this is the matching mechanism that allows for PTM selection. Finally, we introduce how \SeqFusion can benefit from the fusion of multiple most suitable PTMs for forecasting. The overall architecture of \SeqFusion is shown in Figure~\ref{fig:method}.

\subsection{Leveraging PTMs for Time-Series Forecasting}
\label{Section4-1}

In \SeqFusion, we propose to collect diverse PTMs and fuse their sequential predictions for multivariate zero-shot forecasting. Instead of collecting a large amount of diverse data, we create a model zoo containing PTMs pre-trained on various tasks. Specifically, we collect a model zoo $\mathcal{M}= \left\{ \mphi_m \right\}_{m=1}^M$ containing diverse PTMs, where each PTM $\mphi_m:\mathbb{R}^{T \times 1} \rightarrow \mathbb{R}^{H \times 1}$ is an one-variate model that pre-trained on a different dataset $\mathbf{X}_{\mphi_m}$. Given a PTM $\mphi_m$, we can directly predict the future value $\hat{\mathbf{y}}_c = \mphi_m(\mathbf{x}_c)$ based on the corresponding variate of its history $\mathbf{x}_c \in \mathbf{X}$ once the input length $T$ is consistent.

Once the model zoo $\mathcal{M}$ is created, \SeqFusion selects the most suitable PTMs for each variate of the target time series $\mathbf{X}$. We will discuss this selection process later. With the selected PTMs $\{\hat{\mphi}_c\}_{c=1}^C$ for each variate of the target time series $\mathbf{X}$, the forecasting process is executed through the following stages as shown in Figure~\ref{fig:method_b}:\looseness=-1

\noindent{\bf Normalization}. Each variate $\mathbf{x}_c$ of the input time series $\mathbf{X}$ is first normalized to ensure uniformity in data scale and distribution: $\mathbf{x}_c^{\text{norm}} = (\mathbf{x}_c - \mu_{\mathbf{x}_c}) / \sigma_{\mathbf{x}_c}$, where $\mu_{\mathbf{x}_c}$ and $\sigma_{\mathbf{x}_c}$ are the mean and standard deviation of variate $\mathbf{x}_c$. Note that the normalization module decreases the distributional discrepancy among each input variate, making the distribution of the model input more stable.

\noindent{\bf Sequential Fusion}. Following normalization, the process employs a sequential forecasting loop, where the selected pre-trained model $\hat{\mphi}_c$ is applied repeatedly over $n= H/h$ times for each variate with the initial input $\mathbf{x}_c^{\text{norm}}$, and each block involves a trim operation that refines the output time series for the next iteration in Alg.~\ref{al:1}. From here, we get the predictions $\hat{\mathbf{y}}_c^{\text{norm}}$ of every variate.

\noindent{\bf Concatenation and De-normalization}. The final outputs from the sequential forecasting steps are concatenated and then de-normalized to convert them back to the original scale of the data: 
\begin{equation}
 \hat{\mathbf{Y}}=\{\hat{\mathbf{y}}_{1}, ..., \hat{\mathbf{y}}_{C}\}, \ \ \hat{\mathbf{y}}_c = \sigma_{\mathbf{x}_c}\cdot \hat{\mathbf{y}}_c^{\text{norm}} + \mu_{\mathbf{x}_c}.
\end{equation}

The final output of \SeqFusion, $\hat{\mathbf{Y}}$, is a collection of multivariate sequences, each of which represents the predictions for a variate within the specified prediction horizon $H$. This approach not only takes advantage of the PTMs but also implements variate-length prediction through a recursive strategy that can be used to predict any number of independent variates.\looseness=-1

\subsection{Universal Representations for Selection}
\label{Section4-2}
In \SeqFusion, We select suitable PTMs for the target time series based on the representations of target time series and PTMs. We encode the target time series and PTMs using a general extractor model $\psi$. The model $\psi$ is well-trained on a general dataset $\sD$ that does not contain overlapping data with the target time series $\mathbf{X}$. We constructed this generalized dataset $\sD$ by randomly sampling sample sub dataset $\sD_i$ from the pre-training datasets $\{\mathbf{X}_{\mphi_m}\}_{m=1}^M$ of the model zoo. And we extract the representation of $\mathbf{X}_{\mphi_m}$ to characterize each PTM $\mphi_m$ in the model zoo.

\begin{figure*}[t]
\centering
\hspace{-0.2in}
\subfloat[Architecture of General Extractor.]{\includegraphics[width=2.7in]{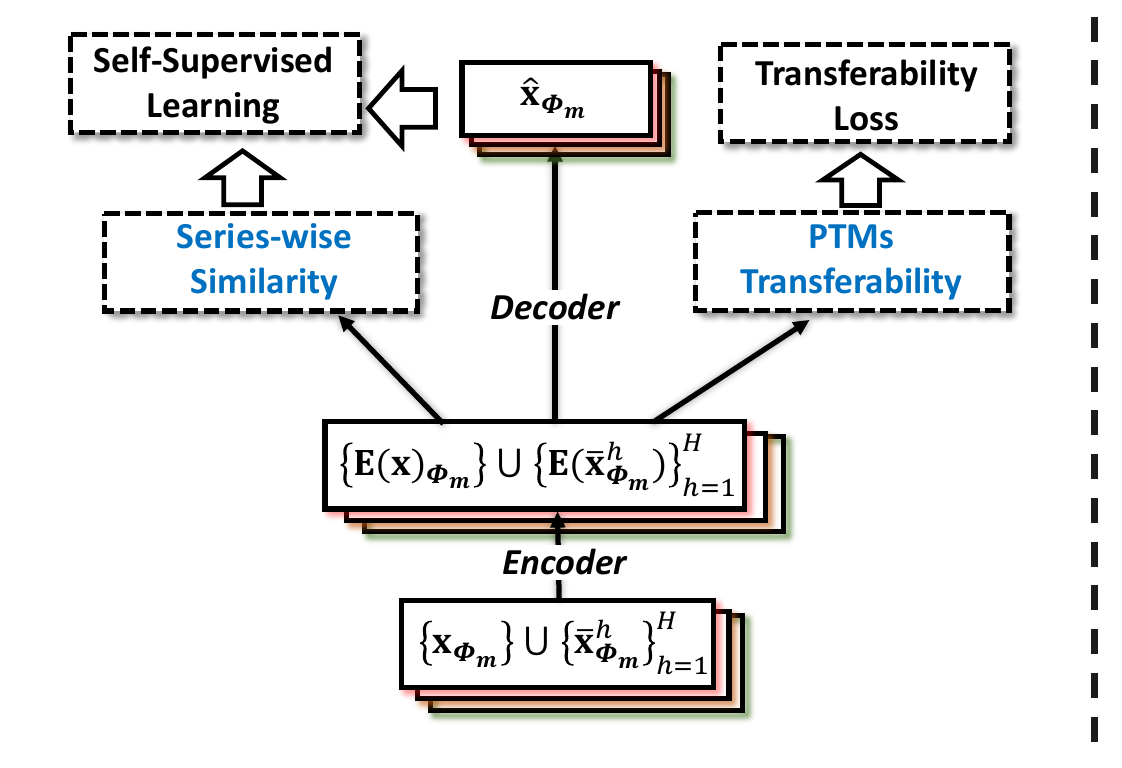}}
\hspace{0.2in}
\subfloat[Training process of General Extractor.]{\includegraphics[width=3.7in]{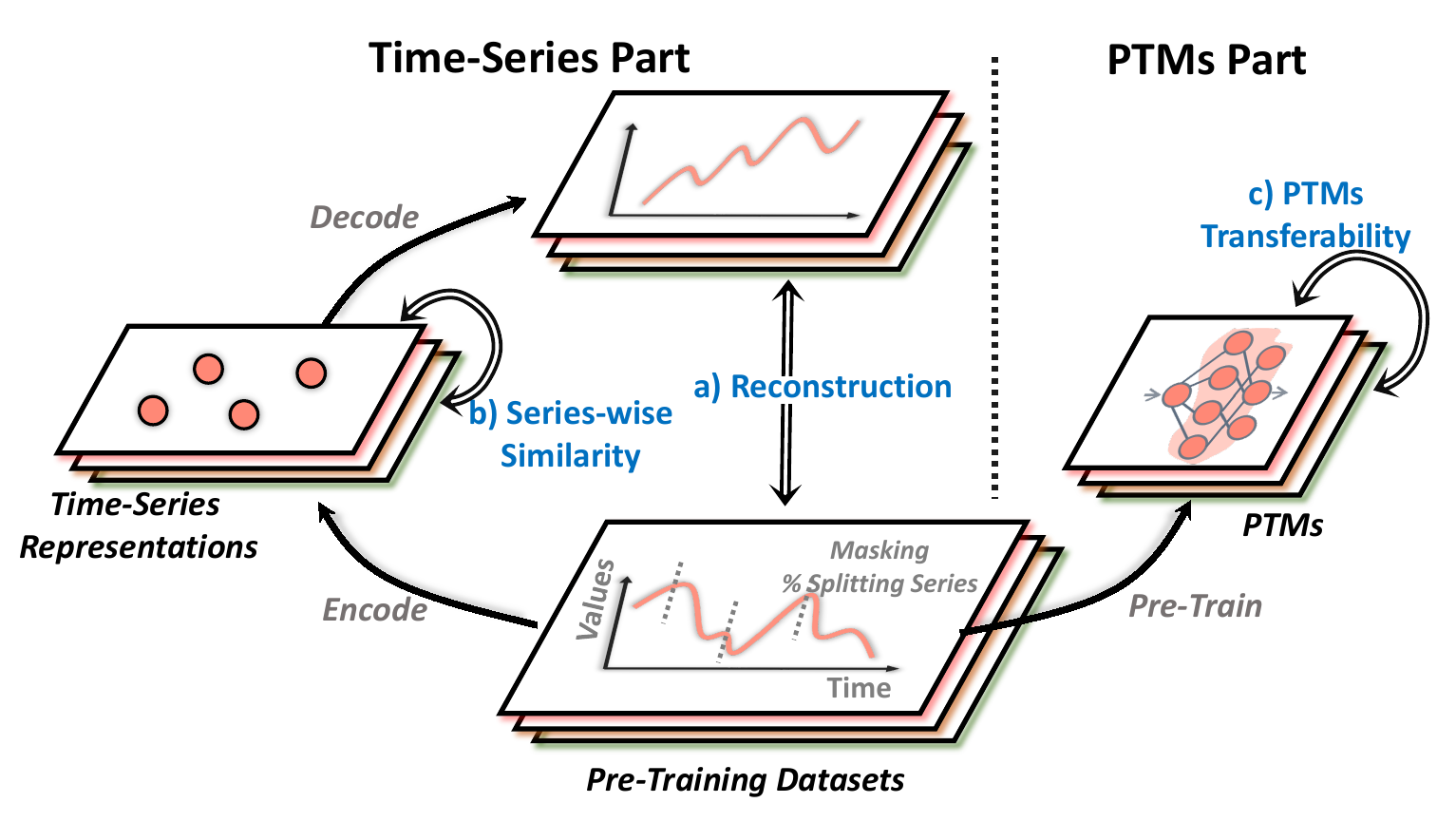}}
\hspace{-0.2in}
\caption{(a) Architecture of the General Extractor. The encoder-decoder model extracts time-series representations optimized through self-supervised learning based on series-wise similarity and transferability loss to align time-series representations with PTMs' transferability. (b) Training Process of the General Extractor. In the Time-Series part, raw time-series data from pre-training datasets are masked and split to create input pairs. The encoder generates time-series representations while the decoder reconstructs the input. A series-wise similarity loss aligns representations of series from the same dataset and allows representations from different datasets to be far apart. In the PTMs Part, PTMs are evaluated for transferability using metrics like 1-MSE, and this information is integrated to refine representations, ensuring they reflect both the intrinsic characteristics of the time series and the ability of PTMs to generalize to related tasks.\looseness=-1}
\label{fig:extractor}
\end{figure*}

\noindent{\bf General Extractor.} The general extractor model $\psi$ is trained using an unsupervised learning process inspired by~\cite{simmtm}, employing an encoder-decoder architecture. The encoder processes input time series to a latent representation $\mathbf{E}(\cdot)$, while the decoder reconstructs the original time series from this latent representation. The output of the encoder, $\mathbf{E}(\cdot)$, is utilized as the representation of the input sequence. 

To further enhance the quality of the representations, as shown in Figure~\ref{fig:extractor}, the training process includes two key components. Firstly, a series-wise similarity objective aligns representations of time series sampled from the same dataset, ensuring consistency within related data. Given a set of input series $\mathbf{x}_{\mphi_{m}}$ (here we use PTMs' pre-training datasets), we generate multiple masked time series set $\overline{\mathbf{x}}_{\mphi_{m}}$ by randomly masking a portion of time points along the temporal dimension. The original time series and its masked series will present close representations and be far away from the representations from other series, where we use $\operatorname{sim}(\cdot, \cdot)$ to denote the cosine similarity:
\begin{equation}
\mathcal{L}_{\text{constraint}} = - \sum_{s \in {\mathbf{x}_{\mphi_{m}}}} \left( \sum_{s' \in \overline{\mathbf{x}}_{\mphi_{m}}} \log \frac{\exp(\text{sim}(s, s'))}{\sum_{s'' \in {\mathbf{x}_{\mphi_{m}}} } \exp(\text{sim}(s, s''))} \right).
\end{equation}

Secondly, the transferability loss aligns the learned representations with the transferability potential of PTMs. Representations learned through reconstruction are inherently time-series-specific and is model-independent, whereas the ability of different PTMs to work on the same training set is likely to be different. in order to better fit the PTM selection mechanism, we introduce the \textit{transferability loss} function. Specifically, given two series $\mathbf{x}_i \in \sD_i$, $\mathbf{x}_j \in \sD_j$, the model $\psi$ is trained with an additional supervised loss:
\begin{equation}
\mathcal{L}_{\text{trans}} = \sum_{\mathbf{x}_i, \mathbf{x}_j \in \sD} \left\| g_{i,j} - \text{sim}\big(\mathbf{E}(\mathbf{x}_i),\mathbf{E}(\mathbf{x}_j)\big) \right\|_2^2,
\end{equation}
where $g_{i,j}$ is the transferability from dataset $\sD_i$ to dataset $\sD_j$. In our experiments, the $g_{i,j}$ is computed using the 1-MSE metric, which measures the MSE metric performance of PTMs trained on $\sD_i$ and tested on $\sD_i$. By aligning the similarity of representations with transferability scores, the extractor $\psi$ learns representations that not only capture time-series features but also reflect the potential utility of PTMs across datasets. Finally, the overall optimization process of the general extractor can be represented as follows:
\begin{equation}
\min_{\Theta} \left\| \mathbf{x}_{\mphi_{m}} - \hat{\mathbf{x}}_{\mphi_{m}} \right\|_2^2 + \mathcal{L}_{\text{trans}} + \lambda \mathcal{L}_{\text{constraint}} ,
\end{equation}
where the first part is the reconstruction loss, $\hat{\mathbf{x}}_{\mphi_{m}}$ denotes the reconstructed output of the decoder, and ${\Theta}$ denotes the set of all parameters of the general extractor. This design ensures that the representations of PTMs pre-trained on related datasets are closely aligned in the embedding space, while representations of dissimilar datasets remain distinct. Consequently, models trained on datasets that are similar to the target time series are assigned higher similarity scores (see Figure~\ref{fig:repr} for more details), facilitating accurate and efficient PTM selection. \looseness=-1

\noindent{\bf Representation Extraction.} 
We extract the variate-wise representations the target time-series $\mathbf{X}$ and the representations of PTMs $\left\{ \mphi_m \right\}_{m=1}^M$ via the general extractor $\psi$. Specifically, for each variates $\mathbf{x}_c $ of the target time series $\mathbf{X}$, we encode them separately, denoted by $\mmu_c \in \mathbb{R}^d$; for each PTM $\mphi_m$, we sampled a batch of data $\mphi_{m}$ from its pre-training dataset $\mathbf{X}_{\mphi_m}$, and use the average encoding $\mtheta_m \in \mathbb{R}^d$ of the sampled data as the representation: 
\begin{equation}
\mmu_c = \psi(\mathbf{x}_c) \big|_{c=1}^C, \quad \mtheta_m = {\psi(\mathbf{x}_{\mphi_{m}})} \big|_{m=1}^M.
\end{equation}
With this form of representation, we project the representation of the model and the representation of the target time series into the same space. Models trained from related datasets are similar in this representation, and models with pre-trained datasets that are similar to the target time series data are similar in this representation. This crucial step makes our subsequent model-matching process feasible.

\subsection{Matching PTMs for Target Time Series}
\label{Section4-3}
We extract the variate-wise representations $\{\mmu_c\}_{c=1}^C$ of the target time series $\mathbf{X}$ and the representations $\{\mtheta_m\}_{m=1}^M$ of PTMs $\left\{ \mphi_m \right\}_{m=1}^M$ via the general extractor $\psi$. We then select PTMs for each variate of $\mathbf{X}$ based on the similarity between the target time-series representation $\{\mmu_c\}_{c=1}^C$ and those $M$ PTMs' representations $\{\mtheta_m\}_{m=1}^M$. We expect that the higher the similarity, the more helpful a PTM is for the target time-series. We use the cosine similarity $\operatorname{sim}(\cdot, \cdot)$ for measuring the similarity, and $\{ \hat{\mphi}_c \}_{c=1}^C$ to represent the selected PTMs for the target time-series with $C$ variates:
\begin{equation}
\label{eq:3}
 \hat{\mphi}_c  = \argmaxB_{{\mphi}_i \in \mathcal{M}} \operatorname{sim}\big(\mtheta_i, \mmu_c\big), \qquad c = 1,...,C.
\end{equation}
Given the target time-series $\mathbf{X}$, we now select a PTM set $\{\hat{\mphi}_c\}_{c=1}^C$ for each variate using Equation~\ref{eq:3}. This selection is based on the historical data and specific characteristics of each variate, ensuring that each model is optimized for the type of data it will process.\looseness=-1

Since each PTM in the model zoo specializes in a different temporal pattern and prediction task, so given any new time series task, \SeqFusion has the opportunity to select it to help with zero-shot prediction, as long as there is a PTM that has been trained on a similar task.

\subsection{Optional Aggregated Predictions}
\label{Section4-4}
To further enhance the accuracy and robustness of the forecasts, \SeqFusion includes an optional aggregated prediction phase. This phase integrates the predictions of $k$ PTMs, selected based on their similarity to the input variate $\mathbf{x}_c$ from Equation~\ref{eq:3}. Specifically, in each block of the sequential forecasting, we give an average prediction based on the top-$k$ suitable PTMs: $\hat{\mathbf{y}}_c^{\text{aggr}}= \sum_{i=1}^k \hat{\mathbf{y}}_{c, i}^{\text{norm}} / k$, where we use $\hat{\mathbf{y}}_{c,i}^{\text{norm}}$ to represent the prediction from one PTM, and the final predictions are:
\begin{equation}
 \hat{\mathbf{Y}}_{\text{aggr}}=\{\hat{\mathbf{y}}_{1}^{\text{aggr}}, ..., \hat{\mathbf{y}}_{C}^{\text{aggr}}\}, \\ \ \hat{\mathbf{y}}_c^{\text{aggr}} = \sigma_{\mathbf{x}_c} \cdot \hat{\mathbf{y}}_c^{\text{aggr}} + \mu_{\mathbf{x}_c}.
\end{equation}
Incorporating aggregated predictions into the \SeqFusion framework enhances the reliability and accuracy of the final forecast (see Figure~\ref{fig:ensemble}). This phase ensures that the model can leverage multiple sources of knowledge, making it more robust in various forecasting scenarios.

% \subsection{Dynamic Replacement of Recursive Blocks}

% The %说明一下时间序列的预测有可能发生动态的变化。使用同一个预训练模型进行recursive的预测可能会导致

% To enhance adaptability, a dynamic replacement step can be included in \SeqFusion, where the selected pre-trained model is updated in response to new data or shifts in data characteristics. This is given by:
% \begin{equation}
% \label{eq:2}
%  \hat{\mphi}_{c}^{t+1} = \mathop{\arg\max}_{{\mphi}_i \in \mathcal{M}} \operatorname{sim}\big(\mtheta_i, \hat{\mathbf{y}}^t\big), \qquad t = 2,...,n,
% \end{equation}
% where we update the model $\hat{\mphi}_{c}^{t+1}$ at the $(t+1)^{th}$ block based on the new observations $\hat{\mathbf{y}}_c^t$ in the recursive process, improving the model's responsiveness to changing data trends. Through this dynamic replacement, \SeqFusion delivers a flexible solution for forecasting tasks across varied domains, especially where traditional forecasting methods may struggle to adapt quickly to new data without extensive retraining.

\section{Experiments}
% 统一交代Dataset; task split;Implementation Details(training and inference);metrics;Baselines
% results: sequential

We evaluate \SeqFusion on three zero-shot benchmarks: multivariate datasets with a) lightweight PTMs or b) large language models, and c) univariate datasets with lightweight PTMs. We then analyze the influence of key components in \SeqFusion.

\subsection{Evaluation on Multivariate Time Series}
\label{Section5-1}
\noindent{\bf Setups}. We follow the setting in previous works~\cite{forecastpfn, Wu2021Autoformer} and evaluate models on seven popular, real-world datasets across various domains, which are summarized in Table~\ref{tb:multi_dataset}. The datasets include: 
\begin{itemize}
	\item \textbf{ETT (Electricity Transformer Temperature)}~\cite{Zhou2021informer}\footnote{\url{https://github.com/zhouhaoyi/ETDataset}}: This dataset comprises two hourly-level datasets (ETTh) and two 15-minute-level datasets (ETTm). Each dataset contains seven oil and load features of electricity transformers from July $2016$ to July $2018$.
	\item \textbf{Traffic}\footnote{\url{http://pems.dot.ca.gov}}: This dataset describes road occupancy rates and contains hourly data recorded by sensors on San Francisco freeways from 2015 to 2016.
	\item \textbf{Electricity}\footnote{\url{https://archive.ics.uci.edu/ml/datasets/ElectricityLoadDiagrams20112014}}: This dataset collects the hourly electricity consumption of $321$ clients from $2012$ to $2014$.
	\item \textbf{Exchange-Rate}~\cite{lai2018modeling}\footnote{\url{https://github.com/laiguokun/multivariate-time-series-data}}: This dataset collects the daily exchange rates of $8$ countries from $1990$ to $2016$.
	\item \textbf{Weather}\footnote{\url{https://www.bgc-jena.mpg.de/wetter/}}: This dataset includes $21$ indicators of weather, such as air temperature, and humidity, recorded every $10$ minutes for year $2020$ in Germany.
	\item \textbf{ILI}\footnote{\url{https://gis.cdc.gov/grasp/fluview/fluportaldashboard.html}}: This dataset describes the ratio of patients seen with influenza-like illness to the total number of patients. It includes weekly data from the Centers for Disease Control and Prevention of the United States from $2002$ to $2021$.
\end{itemize}
% ECL\footnote{\url{https://archive.ics.uci.edu/ml/datasets/ElectricityLoadDiagrams20112014}}, ETTh1~\cite{Zhou2021informer}\footnote{\url{https://github.com/zhouhaoyi/ETDataset}}, ETTh2~\cite{Zhou2021informer}, Exchange~\cite{lai2018modeling}\footnote{\url{https://github.com/laiguokun/multivariate-time-series-data}}, Illness\footnote{\url{https://gis.cdc.gov/grasp/fluview/fluportaldashboard.html}}, Traffic\footnote{\url{http://pems.dot.ca.gov}}, and Weather\footnote{\url{https://www.bgc-jena.mpg.de/wetter/}}. 
Following previous work~\cite{forecastpfn}, the look-back window length is set to 36 and the prediction horizon is set to $\{6, 8, 14, 18, 24, 36, 48\}$ for all datasets. We measure the performance using Mean Squared Error (MSE).
    \begin{equation}
		\setlength\abovedisplayskip{3pt}
		\setlength\belowdisplayskip{3pt}
    \text{MSE}= \frac{1}{H}\sum_{i=1}^H (\mathbf{Y}_i - \hat{\mathbf{Y}}_i)^2
\end{equation}
where we assume that $\mathbf{Y}, \hat{\mathbf{Y}} \in \mathbb{R}^{H \times C}$ are the ground truth and prediction results of the future with $H$ time points and $C$ variates. $\mathbf{Y}_i$ denotes the $i$-th future time point.

\begin{table}[t]
\centering
\setlength{\tabcolsep}{0.9mm} % Adjust the column separation
  \begin{tabular}{ccccc}
    \toprule
    \textbf{Dataset} & \textbf{Channels}  & \textbf{Timesteps} &\textbf{Granularity} & \textbf{Domain} \\
    \toprule
     ETTh1, ETTh2 & 7  & 17,420 & Hourly & Electricity\\
     % \midrule
     % ETTm1, ETTm2 & 7  & 69,680 & 15min & Electricity\\
     \midrule
     Exchange-Rate & 8 & 7,588  & Daily & Economy\\
    \midrule
    Weather & 21  & 52,696  & 10min & Weather\\
    \midrule
    ECL & 321  & 26,304 & Hourly & Electricity \\
    \midrule
    Traffic & 862 & 17,544 & Hourly & Transport \\
    \midrule
    ILI  & 7 & 966  & Weekly & Health \\
    \bottomrule
    \end{tabular}
    \vspace{5pt} 
  \caption{Statistics of multivariate time series datasets.}\label{tb:multi_dataset}
  \vspace{-15pt} 
\end{table}

% \begin{figure*}[t]
%     \begin{subfigure}[b]{0.54\linewidth}
%         \centering
%         \includegraphics[width=\linewidth]{fig/repr.pdf}
%          \caption{PCA plot of PTM repr. and target datasets repr.}
%          \label{fig:repr}
%     \end{subfigure}
%     \begin{subfigure}[b]{0.41\linewidth}
%     \centering
%         \includegraphics[width=\linewidth]{fig/zoo.pdf}
%          \caption{MSE performance distribution of all PTMs.}
%          \label{fig:zoo}
%     \end{subfigure}
%   \caption{(a) Visualization of the representations (repr.) of PTMs and target time series. We use similar colors to indicate different variate of the same time series. Notice that some variates with the same sampling frequency are cluster closer. (b) Violin plot showing the ground truth performance distribution of PTMs on target datasets. The red ``x'' markers indicate the performance of \SeqFusion. Markers near the bottom indicates that \SeqFusion selects those most suitable PTMs.}
% \end{figure*}

\begin{table*}[ht]
\centering
\setlength{\tabcolsep}{1.25mm} % Adjust the column separation
\begin{tabular}{lccccccccc}
\toprule
\multirow{2}{*}{\textbf{Methods}} & \textbf{Resource} &\multicolumn{7}{c}{\textbf{Downstream Target Dataset}} & {\textbf{Memory Storage (MB)}} \\
& \textbf{Type}&
  \multicolumn{1}{l}{ECL} &
  \multicolumn{1}{l}{ETTh1} &
  \multicolumn{1}{l}{ETTh2} &
  \multicolumn{1}{l}{Exchange} &
  \multicolumn{1}{l}{Illness} &
  \multicolumn{1}{l}{Traffic} &
  \multicolumn{1}{l}{Weather} & Data + Model\\
\midrule
Last        &  \multirow{3}{*}{-}& 0.7360  & 0.7640 & \underline{0.2639}  & \underline{\textbf{0.0217}}   & 4.7867 & 2.2498 & 1.4799 & \multirow{3}{*}{-} \\
Mean       &   & 0.6755  & \underline{0.6314} & 0.3550  & 0.0376   & 4.8981 & {1.3565} & \textbf{\underline{{1.4063}}} &\\
SeasonalNaive && {0.6091} & 0.8539 & 0.3315  & 0.0272   & 6.0760 & {1.2227} & 1.6105 & \\
\midrule
Arima       & \textbf{In-Task}& 3.6648  & 0.6389 & 1.0048  & 10.1624  & 5.8628 & 2.4790 & 3.1264 & 0.01 + 30.27\\
Prophet     & \textbf{Data} & 10.2358 & 6.1366 & 10.1787 & 229.8594 & 9.1147 & 4.8610 & 2.9049 & 0.01 + 3.270\\
\midrule
Transformer &  & 1.3429  & 0.6875 & 0.9453  & 1.5532   & 5.0552 & 1.9336 & 2.1727 & 0.01 + 64.06\\
Autoformer   & & 0.8861  & 0.8519 & 0.5835  & 0.1950   & \underline{4.5547} & 1.4316 & 1.7660 & 0.01 + 65.88\\
FEDformer   & \multirow{2}{*}{\textbf{In-Task}} & 0.9156  & 0.7561 & 0.4718  & 0.0943   & 4.6087 & 1.5551 & 1.6792&  0.01 + 80.38\\
Informer    & \multirow{2}{*}{\textbf{Data}} & 1.3743  & 0.7870 & 0.8497  & 1.5969   & 5.3082 & 2.0612 & 2.3070 & 0.01 + 67.07\\
DLinear     &  & 0.6942  & 0.6472 & 0.3730  & 0.0559   & 4.8826 & 1.3655 & 1.4644 & 0.01 + 0.050\\
PatchTST   &   & 0.6184 & 0.7333 & 0.4006 & 0.0354 & 3.9034 & \underline{1.1661} & 1.4877& 0.01 + 2.310\\
iTransformer  &   &\underline{0.6067} & 0.7183 & 0.3345 & 0.0315 & 3.5232 & \textbf{\underline{{1.1306}}} & 1.5676 & 0.01 + 26.15\\
\midrule
Meta-N-BEATS & \multirow{2}{*}{\textbf{Pre-Train}}  & 0.7576  & 0.7715 & 0.3133  & 0.0469   & 4.6405 & 2.2361 & 1.4648 & 1.70 + 95.85\\
GPT4TS     & \multirow{2}{*}{\textbf{Data}}  & 0.7458  & 0.6961 & 0.3397  & 0.0280   & 6.9384 & 1.6730 & 1.4777 & 1.70 + 314.83\\

ForecastPFN  & & 0.9511  & 1.1851 & 0.5144  & 0.0579   & 4.8880 & 1.7894 & 1.8770 & * + 23.50\\

% Chronos & & \textbf{0.5444}  &	\textbf{0.5652} &	\textbf{0.2323}  & 0.0239   & 5.7046 & \textbf{1.0351} & \textbf{1.3580} & 7.92*10$^5$ + 242.13\\   	 	 	 	 
\midrule  		 				
\SeqFusion  & \textbf{PTMs} & \textbf{\underline{{0.6029}}} & \textbf{\underline{{0.6001}}} & \textbf{\underline{{0.2450}}}& \underline{\textbf{0.0217}} & \underline{\textbf{3.4956}} & 1.4889 &\underline{1.4488} & 0.02 + 23.10\\
\bottomrule
\end{tabular}
\vspace{5pt} 
\caption{Performance comparisons on multivariate forecasting tasks over 5 trials, including classical statistical methods, deep learning models trained on 50 in-task time steps data, and zero-shot methods requiring pre-training data. \SeqFusion achieves competitive performance across most datasets while requiring minimal memory storage. We denote the \underline{\textbf{best}} and \underline{second-best} results with bold-underline and underline. $^{*}$Synthetic datasets.\looseness=-1} 
  \label{tab:multivariate}
  \vspace{-10pt}
\end{table*}

\begin{table*}[ht]
\centering
\setlength{\tabcolsep}{1.25mm} 
\begin{tabular}{lccccccccc}
\toprule
\multirow{2}{*}{\textbf{Methods}} & \textbf{Resource} &\multicolumn{7}{c}{\textbf{Downstream Target Dataset}} & {\textbf{Memory Storage (MB)}} \\
& \textbf{Type}&
  \multicolumn{1}{l}{ECL} &
  \multicolumn{1}{l}{ETTh1} &
  \multicolumn{1}{l}{ETTh2} &
  \multicolumn{1}{l}{Exchange} &
  \multicolumn{1}{l}{Illness} &
  \multicolumn{1}{l}{Traffic} &
  \multicolumn{1}{l}{Weather} & Data + Model\\
\midrule

TimeGPT$^*$    &   & 0.7458  & 0.6961 & 0.3397  & 0.0280   & 6.9384 & 1.6730 & 1.4777 & $*$\\
Chronos & \textbf{Pre-Train} & \underline{0.5444}  &	\underline{0.5652} &	\underline{\textbf{0.2323}}  & \underline{\textbf{0.0239}}   & 5.7046 & {1.0351} & \underline{1.3580} & 7.92$\times$10$^5$ + 242.13\\     
% Moirai & \textbf{Data}& 0.6071 & 0.5978& 0.2596& 0.0224 & 8.0574 & 1.0086 & 1.3143 & 3.56$\times$10$^5$ + 365.15\\
Moirai & \textbf{Data}& 0.6909 & 0.6986& 0.3102& 0.0372 & {5.2899} & 1.4818 & 1.5127 & 3.56$\times$10$^5$ + 365.15\\
TimesFM & & 0.6100 & 0.5671 & 0.4721 & 0.0355 & \underline{\textbf{3.1143 }}& \underline{\textbf{0.8722}} & 1.4332 & $\ge$ 1.00$\times$10$^6$ + 814.76\\
\midrule  		 				
\SeqFusion  & \textbf{PTMs} & \underline{\textbf{0.5263}} & \underline{\textbf{0.5265}}&\underline{0.2604}&\underline{0.0244}&\underline{4.3535}&\underline{0.8775} & \underline{\textbf{1.3323}}&0.02 + 1.42$\times$10$^3$\\

\bottomrule
\end{tabular}
\vspace{5pt} 
\caption{Performance comparisons between large-scale pre-trained models and our \SeqFusion framework, utilizing a large-scale pre-trained model zoo. \SeqFusion achieves competitive or superior performance across most datasets with significantly lower memory storage requirements. We denote the \underline{\textbf{best}} and \underline{second-best} results with bold-underline and underline. $^*$Closed-source. }
\label{tab:llms}
\vspace{-15pt}
\end{table*}

\noindent{\bf Baselines}. 
To provide a comprehensive evaluation, we include a variety of forecasting models:
\begin{itemize}
    \item Naive models: \textbf{Last}, which uses the last observed value for future predictions; \textbf{Mean}, which predicts future values as the mean of past observations; and \textbf{SeasonalNaive}~\cite{Hoptroff93} (with a period of 7), which repeats the values from the same period in the previous cycle.\looseness=-1
    \item Statistical methods: we include \textbf{ARIMA}~\cite{Hoptroff93}, known for its autoregressive integrated moving average properties, and \textbf{Prophet}~\cite{box1970distribution}, a model developed by Facebook designed for handling seasonality and trends in time series data.
    \item Deep learning models: \textbf{Transformer}~\cite{vaswani2017attention}, renowned for its attention mechanism; \textbf{Autoformer}~\cite{Wu2021Autoformer}, which introduces decomposition blocks to enhance the representation of seasonal patterns; \textbf{Informer}~\cite{Zhou2021informer}, optimized for efficiency and scalability in handling large datasets; \textbf{DLinear}~\cite{DLinear}, focusing on decomposed linear forecasting; \textbf{PatchTST}~\cite{patchtst}, which leverages patch-based temporal patterns; and \textbf{iTransformer}~\cite{iTransformer}, an inverted Transformer designed for multivariate time series.\looseness=-1
    \item Zero-shot forecasting models: \textbf{Meta-N-BEATS}~\cite{metanbeats}, which adapts the N-BEATS architecture for meta-learning; \textbf{ForecastPFN}~\cite{forecastpfn}, a synthetically-trained zero-shot forecastor; and \textbf{GPT4TS}~\cite{onefitsall}, a variant of GPT-2 fine-tuned for time series forecasting. We use Meta-N-BEATS and GPT4TS pre-trained on M4 for multivariate forecasting bennchmark. 
\end{itemize}
The deep learning methods are allowed to train on $50$ in-task time stamps as training data, even though this is not a fair comparison. Others see only the input of length 36 and then predict the future. We reproduce all the deep learning methods and zero-shot ones using the officially provided codes.\looseness=-1
 
% We compare to a variety of established models, spanning from traditional baselines (ARIMA~\cite{Hoptroff93} and Prophet~\cite{box1970distribution}) to deep learning baselines (Transformer~\cite{vaswani2017attention}, Autoformer~\cite{Wu2021Autoformer}, FEDformer~\cite{fedformer}, Informer~\cite{Zhou2021informer}, DLinear~\cite{DLinear}, PatchTST~\cite{patchtst} and iTransformer~\cite{iTransformer}). We also compare three numerical baselines (Last, Mean, and SeasonalNaive) and three state-of-the-art zero-shot baselines (Meta-N-BEATS~\cite{metanbeats}, ForecastPFN~\cite{forecastpfn}
% % Chronos~\cite{chronos} 
% and GPT4TS~\cite{onefitsall}). TimeGPT~\cite{TimeGPT} is closed-source and excluded from our comparison. 

\noindent{\bf Implementation Details}. Our focus is on how PTMs can be utilized to help with downstream time series forecasting tasks. To demonstrate the superiority of our approach, we first collect several super-lightweight PTMs (only around 0.05 MB to 2.31 MB per PTMs), to ensure the memory cost of these models (DLinear and PatchTST) are much smaller than storing the pre-training datasets. Specifically, for multivariate forecasting tasks, we collect 10 PTMs pre-trained on the random sampling set from M3 (3 PTMs), M4 (5 PTMs) and Tourism (2 PTMs) based on the PatchTST~\cite{patchtst} architecture. These datasets are sampled at frequencies ranging from minutes to months, with a variety of domains and covering frequencies common to time series. We set the input length to 36, the predict length to 12, the patch length to 16, the encoder layer to 1, the model dimension to 64, and train with MSE loss for 10 epochs with learning rate of 0.001. For the general extractor model, we train an encoder-decoder model~\cite{simmtm} as the general extractor based on the  pre-training datasets of our PTMs to avoid data leakage. During training, we add a transferability loss to help representations extraction as mentioned above. Specifically, we randomly sample about 300,000 sub series from M3, M4 and Tourism. We conduct a cross-dataset evaluation to get the MSE performance of our model zoo, and use 1 minus this MSE performance (1- MSE) as the ground truth of the transferability loss. Since \SeqFusion does not need training, once the model zoo is constructed, we can directly apply it for evaluation on target datasets, where we set the aggregated size to $3$.\looseness=-1

%These training regimes are different from the other, non-zero-shot models which are trained for each configuration. we compare the zero-shot methods to other non-zero-shot methods described above following~\cite{forecastpfn}, even though this is not a fair comparison: the non-zero-shot methods are allowed to train on earlier portions of the time series used at test time, as is common in standard forecasting algorithms. We restrict the data budget of the non-zero-shot methods to lengths of $50$.

\noindent{\bf Results}. Table~\ref{tab:multivariate} presents our findings for multivariate time-series forecasting tasks. We see that \SeqFusion achieves the lowest and second lowest MSE value across all datasets compared to other baselines. For zero-shot methods, \SeqFusion outperforms GPT4TS even though GPT4TS uses a GPT-2 model with a much larger number of parameters than our approach. We also find that the performance of naive baselines, Last, Mean, and SeasonalNaive depends on the dataset, being strong on some datasets and very weak at other times. Our method, on the other hand, shows consistent performance on all datasets. \looseness=-1

\begin{figure*}[t]
\centering
\hspace{-0.4in}
\subfloat[PCA plot of PTM repr. and target datasets repr.]{\includegraphics[width=3.4in]{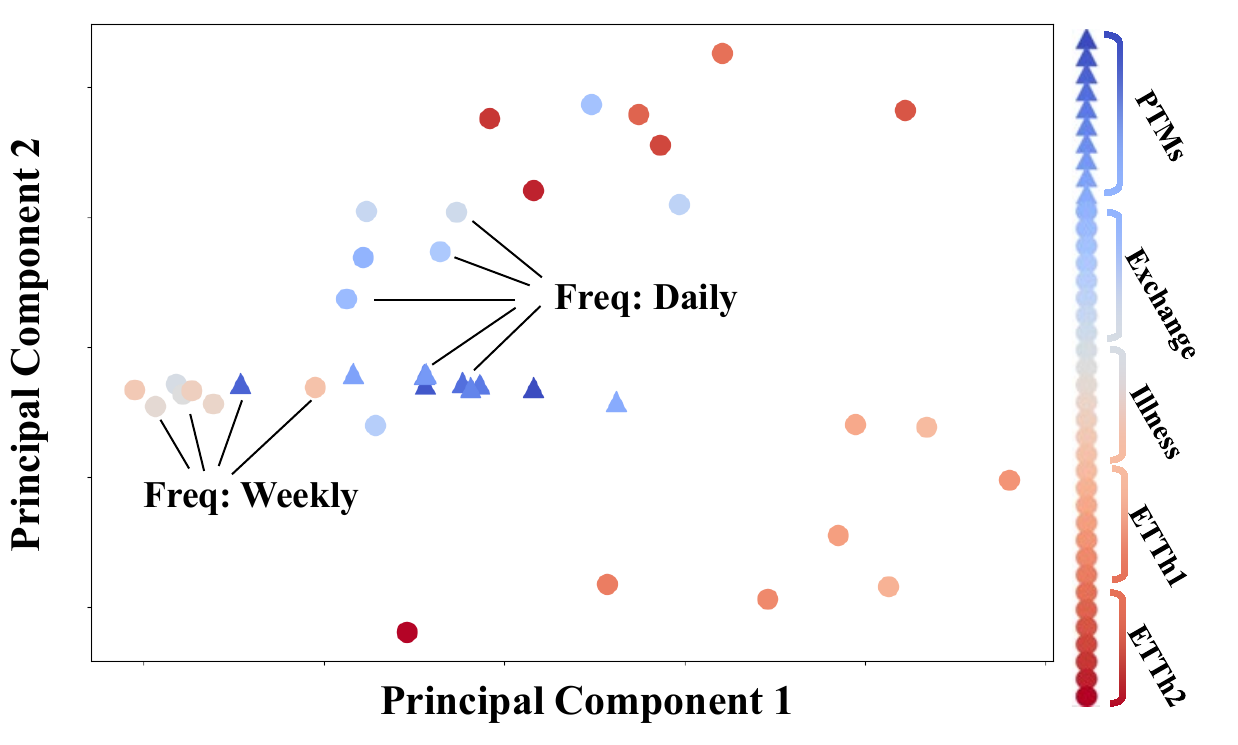}%
\label{fig:repr}}
\hspace{0.3in}
\subfloat[Performance distribution of all PTMs.]{\includegraphics[width=2.8in]{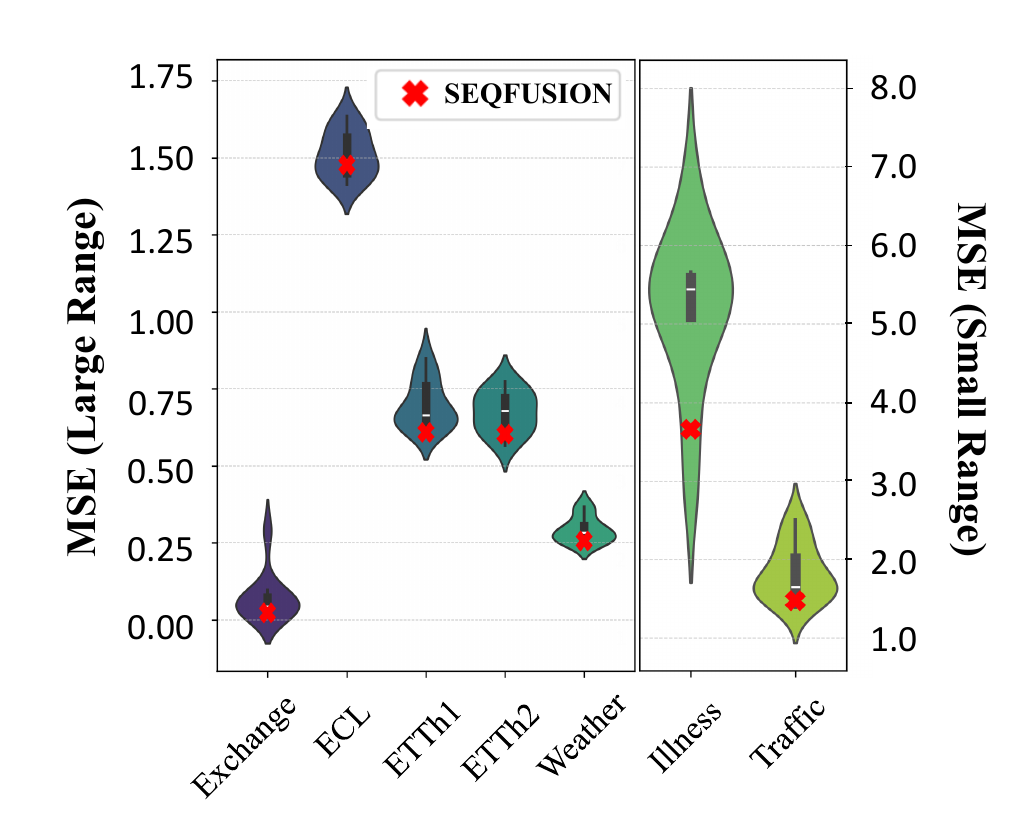}%
\label{fig:zoo}}
\hspace{-0.2in}
\caption{(a) Visualization of the representations (repr.) of PTMs and target time series using PCA. Triangles represent PTMs, and circles represent variates from the target datasets, where we use similar colors to indicate different variate of the time series from one dataset. Variates from the same dataset cluster closely, and datasets with similar sampling frequencies (e.g., daily or weekly) are aligned with PTMs trained on datasets sharing these frequencies. (b)  Violin plot of PTM performance distributions across datasets, with red “x” markers showing \SeqFusion’s combined performance. \SeqFusion consistently selects the most suitable PTMs in the model zoo (markers near the bottom), though its performance is limited by the quality and diversity of the model zoo.}
\end{figure*}

\SeqFusion is able to make good use of the PTM in the model zoo that best fits with the target time series for sequence prediction. As shown in Figure~\ref{fig:zoo}, we plot the distribution of the ground true performance of the PTMs of our model zoo on all datasets, with the red ``x'' representing the combined performance of  the selected PTMs with \SeqFusion. It can be seen that in the vast majority of the datasets, \SeqFusion selected a batch of the most suitable PTMs (box plot near the bottom). At the same time, the upper limit of the performance of \SeqFusion is limited by the breadth of models in the model zoo. It can be found that even though \SeqFusion selected the suitable PTMs for the traffic dataset, limited by the optimal PTMs, \SeqFusion's performance is not satisfactory.

\noindent{\bf Visualization}. To further illustrate why \SeqFusion works, we perform some visualizations. First, we extract the representations of PTMs in the model zoo and the variate-wise representations of the target time series. We then apply PCA to reduce the dimensionality of these representations for visualization. As shown in Figure~\ref{fig:repr}, the triangles represent all PTMs and each circle represents one variate of the target time series. We first see that variates from the same dataset are cluster closer. Also, datasets with similar sampling frequencies (e.g., daily or weekly) are closer in the representation space to PTMs trained on datasets with matching frequencies. For instance, the Exchange dataset, which has a daily sampling frequency, is closely aligned with PTMs trained on the M4-Daily dataset. Similarly, datasets with weekly frequencies exhibit proximity to PTMs trained on datasets with similar temporal patterns. This visualization demonstrates that the representations effectively capture intrinsic characteristics of the time series by the general extractor, such as sampling frequency, and align them with PTMs trained on datasets sharing these characteristics. Such alignment is critical for \SeqFusion’s success, as it ensures that the selected PTMs are not only relevant but also well-suited for the target time series.

\subsection{Evaluation on Large Scale Pre-trained Models}
\label{Section5-2}
\noindent{\bf Setups} In the first experiment, we utilized a set of 10 lightweight PTMs, which demonstrates that even with minimal model complexity, our approach could achieve competitive results. To further evaluate \SeqFusion’s scalability and accuracy, we evaluate our method against state-of-the-art methods using pre-trained models, particularly those based on large language models or pre-trained on large-scale time-series datasets.

\noindent{\bf Baselines}. 
We select several baselines represent various approaches to time-series forecasting using pre-trained models, which includes:
\begin{itemize}
    \item \textbf{TimeGPT}~\cite{TimeGPT}: a large-scale model pre-trained on diverse time series datasets. It demonstrates competitive or superior zero-shot forecasting performance when compared to fully supervised models. We use its official API to perform forecasting. 
    \item \textbf{Chronos}~\cite{chronos}: a T5-based model that tokenizes time-series values using scaling and quantization, creating a fixed vocabulary for the data. Chronos employs transformer-based language model architectures, trained via cross-entropy loss, to process tokenized time-series sequences. We choose the official chronos-small version.
    \item \textbf{Moirai}~\cite{MOIRAI}: a masked encoder-based universal time-series forecasting transformer, Moirai is trained on a large-scale datasets, which contains over 27 billion observations across nine domains. We choose the official Moirai-base version.
    \item \textbf{TimesFM}~\cite{TimesFM}: a large foundation model utilizing a patched-decoder attention mechanism, pre-trained on a large time-series corpus to capture temporal dependencies effectively.
\end{itemize}

\noindent{\bf Implementation Details}. We construct 3 open-source pre-trained models to construct our model zoo and deploy the \SeqFusion framework. This involves Chronos~\cite{chronos}, Moirai~\cite{MOIRAI}, and TimeFM~\cite{TimesFM}.These models represent diverse architectures and pre-training approaches, ensuring that our model zoo captures a wide range of temporal patterns and forecasting capabilities. To demonstrate the generalization ability of \SeqFusion, we reuse the general extractor model trained in the first multivariate experiment, avoiding the need for task-specific re-training. To extract model representations, we sample 100,000 time-steps from the large-scale training datasets of Chronos, Moirai, and TimeFM as representative subsets. We assess the zero-shot performance on the multivariate forecasting tasks. 

\noindent{\bf Results} Table~\ref{tab:llms} demonstrates that \SeqFusion achieves competitive performance compared to state-of-the-art large-scale pre-trained models across diverse datasets while maintaining exceptional storage efficiency. For instance, \SeqFusion outperforms all baselines on ECL (\textbf{0.5263 MSE}) and Weather (\textbf{1.3323 MSE}), demonstrating its ability to handle datasets with complex temporal dependencies. Additionally, \SeqFusion provides robust accuracy on other datasets, such as Illness and ETTh2, showcasing its versatility across different forecasting domains.

A key advantage of \SeqFusion is its storage efficiency and its data privacy protection. While Moirai, Chronos, and TimesFM require storage of 3.56$\times$10$^3$ MB, 7.92$\times$10$^3$ MB, and over 1.00$\times$10$^6$ MB, respectively, \SeqFusion only requires 1.42$\times$10$^3$ MB. This reduction in memory requirements makes \SeqFusion a practical choice for resource-limited environments where large-scale models may be impractical to deploy. Moreover, \SeqFusion achieves this efficiency through collecting black-box PTMs with a few its pre-training data (less than 0.1\%), making it particularly suitable for real-world applications that need to protect data privacy.\looseness=-1

The results also validate the scalability of \SeqFusion, demonstrating its effectiveness across different model types, from lightweight to large-scale pre-trained models. By dynamically selecting and aggregating predictions from PTMs in the model zoo, \SeqFusion achieves accuracy comparable to or better than large-scale pre-trained models. This flexibility ensures that \SeqFusion is not only a powerful tool for zero-shot time-series forecasting but also a scalable and resource-efficient framework capable of adapting to a wide variety of forecasting scenarios.\looseness=-1

\subsection{Evaluation on Univariate Time Series}

\begin{table}[t]
\centering
\setlength{\tabcolsep}{0.8mm} % Adjust the column separation
    \begin{tabular}{cccccc}
    \toprule
        \textbf{Dataset} & \textbf{Granularity} & \textbf{Series Size} & \textbf{Look-back} & \textbf{Horizon} & \textbf{Mapping} \\
    \midrule
        \multirow{4}{*}{M3} & Yearly           & 645  & 12 & 6  & M4-Yearly\\
         & Quarterly        & 756  & 24 & 8 & M4-Quarterly\\
         & Monthly          & 1,428  & 24 & 18  & M4-Monthly\\
         & Others           & 174   & 16 & 8 & M4-Monthly\\
        \midrule
        \multirow{6}{*}{M4} & Yearly           & 23,000  & 9 & 6 & M3-Yearly\\
         & Quarterly        & 24,000  & 16 & 8 & M3-Quarterly\\
         & Monthly          & 48,000  & 36 & 18& M3-Monthly\\
         & Weekly           & 359     & 65 & 13& M3-Monthly\\
         & Daily            & 4227    & 9  & 14& M3-Monthly\\
         % & Hourly           & 414     & 2  & 48& M3-Monthly\\  
         
         \midrule
          \multirow{3}{*}{Tourism} & Yearly           & 518  & 12 & 4& M3-Yearly\\
         & Quarterly        & 427  & 24 & 8& M3-Quarterly\\
         & Monthly          & 366  & 36 & 24& M3-Monthly\\
        
    \bottomrule
    \end{tabular}
    \vspace{5pt}
	\caption{Statistics of univariate time series datasets.}
    \label{tb:uni_dataset}
    \vspace{-15pt}
\end{table}

\label{Section5-3}
\noindent{\bf Setups}. For a more comprehensive comparison of performance, we use the univariate time series benchmark from previous works~\cite{metanbeats, onefitsall}, including M3, M4 and Tourism. In this benchmark, we evaluate all baselines in a transfer setting, \ie, how well a model trained from source dataset B performs on target dataset A (without any training data from A).  For M4, Tourism forecasting tasks, we utilize M3 with the mapped granularity as the source dataset. For M3 forecasting tasks, we utilize M4 with the mapped granularity as the source dataset. Following~\cite{onefitsall}, We use the source datasets with the same frequency as the target dataset in these baselines, although it is an unfair comparison to \SeqFusion . We summarize the datasets in Table~\ref{tb:uni_dataset}. We measure the performance using Mean Absolute Percentage Error (MAPE) and Symmetric Mean Absolute Percentage Error (SMAPE):
    \begin{equation}
		\setlength\abovedisplayskip{3pt}
		\setlength\belowdisplayskip{3pt}
    \text{MAPE}= \frac{100}{H}\sum_{i=1}^H \frac{|\mathbf{Y}_i - \hat{\mathbf{Y}}_i|}{|\mathbf{Y}_i| }
    \end{equation}
    \begin{equation}
    		\setlength\abovedisplayskip{3pt}
    		\setlength\belowdisplayskip{3pt}
        \text{SMAPE}= \frac{200}{H}\sum_{i=1}^H \frac{|\mathbf{Y}_i - \hat{\mathbf{Y}}_i|}{|\mathbf{Y}_i| + |\hat{\mathbf{Y}}_i|}
    \end{equation}
where we assume that $\mathbf{Y}, \hat{\mathbf{Y}} \in \mathbb{R}^{H \times 1}$ are the ground truth and prediction results of the future with $H$ time points and $C$ channels. $\mathbf{Y}_i$ denotes the $i$-th future time point.

\noindent{\bf Implementation Details}. We construct the model zoo with 15 models of DLinear architecture pre-trained across traffic, weather, exchange, ETTh1 and ETTh2. We retrain a general extractor based on these pre-training data (randomly sample about 150,000 sub-series) to avoid data leakage from evaluation datasets. We set the aggregated size to $2$. 

\begin{table}[ht]
\centering
\begin{tabular}{lccc}
\toprule
\multirow{2}{*}{\textbf{Methods}} &  \multicolumn{3}{c}{\textbf{Downstream Target Dataset}}   \\

   & M4     & M3       & Tourism \\
    & (SMAPE)       & (SMAPE)       & (MAPE)  \\
\midrule
Last          & 13.1685  & 16.2198  & 35.6551    \\
Mean          & 18.1079  & 21.9992  & 47.3637     \\
SeasonalNaive & 16.5641  & 19.5313  & 40.6277     \\
\midrule
Transformer   & 152.6426 & 117.4006 & 134.8582  \\
Autoformer    & 18.1650  & 17.6839  & 138.1960   \\
Informer      & 152.6291 & 147.6470 & 133.9367  \\
DLinear       & 15.3976  & 17.9384  & 34.9516  \\
PatchTST      & 12.0447  & 13.7123  & \underline{\textbf{25.4164}}    \\
% iTransformer  & 12.13  & 13.93  & 26.91  & 17.66   \\
\midrule
Meta-N-BEATS       & 11.9463  & 13.3581  & \underline{27.3988}    \\
ForecastPFN   & 18.1150  & 21.9990  & 47.3637    \\
GPT4TS        & \underline{\textbf{10.4272}}  & \underline{\textbf{11.6235}}  & 33.7843   \\
\midrule
\SeqFusion     & \underline{11.1604}  & \underline{12.8145}  & 30.4334      \\
\bottomrule
\end{tabular}
\vspace{5pt} 
\caption{Performance comparisons on univariate forecasting tasks over 5 trials. All methods except \SeqFusion and naive baselines use training datasets with the mapped frequency of the target dataset.}
\label{tab:univariate}
\vspace{-15pt}
\end{table}

\noindent{\bf Results}. Table~\ref{tab:univariate} presents the results for univariate time series. We observe that while GPT4TS with the powerful GPT2-backbone (around 314.83MB) achieves the highest rank, \SeqFusion leveraging several simple MLP-based PTMs (only around 0.05MB) demonstrates a great alternative. This shows the importance of leveraging suitable models tailored to specific temporal patterns and forecasting tasks rather than using a general model (We will discuss this later in the ablation studies section). Unlike other deep learning baselines, \SeqFusion does not require training and ranks second with no prior frequency information of the target time series, which highlights its potential for efficient time-series forecasting across various applications in a zero-shot setting.\looseness=-1

\subsection{Ablation Studies}
\label{Section5-4}
We analyze the properties of \SeqFusion following the multivariate benchmark.

% \begin{figure*}[t]
%     \begin{subfigure}[b]{0.53\linewidth}
%         \centering
%         \includegraphics[width=\linewidth]{fig/gt.pdf}
%          \caption{Comparison of the use of transferability loss.}
%          \label{fig:gt}
%     \end{subfigure}
%     \begin{subfigure}[b]{0.53\linewidth}
%     \centering
%         \includegraphics[width=\linewidth]{fig/ensemble.pdf}
%          \caption{Number of Aggregated PTMs vs. MSE.}
%          \label{fig:ensemble}
%     \end{subfigure}
%   \caption{(a) PCA visualizations of PTMs' representations. Without transferability loss (left), PTMs from the same datasets are more dispersed (longer green arrow); With transferability loss (right), PTMs from similar datasets cluster closer together (short green arrow), and vice versa for the red arrow. (b) Performance of \SeqFusion using different numbers of aggregated PTMs across various datasets.}
% \end{figure*}

\noindent{\bf General or Specialized}. The model zoo we constructed consists of PTMs well-trained on diverse datasets, each acquiring \textit{specialized} knowledge. When addressing downstream tasks, \SeqFusion is designed to select the most suitable PTMs for forecasting. To illustrate the effectiveness and necessity of this selection process, we trained a \textit{general} model using PatchTST on the entire pre-training data of our model zoo. Table~\ref{tab:general} demonstrates that the generally pre-trained model performs worse than the selected specialized PTMs. This underscores the importance of leveraging specialized models tailored to specific temporal patterns and forecasting tasks. By selecting PTMs that are finely tuned to particular characteristics of the target time series, \SeqFusion can achieve more accurate and reliable predictions. This approach not only enhances forecasting performance but also mitigates the need for extensive and potentially privacy-compromising data collection. Furthermore, the ability to dynamically choose the most appropriate models from the zoo allows \SeqFusion to adapt to a wide range of forecasting scenarios, making it a versatile and robust solution.

\begin{table*}[t]
\centering
\setlength{\tabcolsep}{1.25mm} % Adjust the column separation

\begin{tabular}{lccccccc}
\toprule
\multirow{2}{*}{\textbf{Methods}} &  \multicolumn{7}{c}{\textbf{Downstream Target Dataset}}   \\
&
  \multicolumn{1}{l}{ECL} &
  \multicolumn{1}{l}{ETTh1} &
  \multicolumn{1}{l}{ETTh2} &
  \multicolumn{1}{l}{Exchange} &
  \multicolumn{1}{l}{illness} &
  \multicolumn{1}{l}{traffic} &
  \multicolumn{1}{l}{weather} \\
\midrule
General Model   & 0.8291 & {0.7723} & 0.3173  & 0.0352   & {5.7027} & {1.6742} & 1.5435\\
Specialized Models            & {\textbf{0.6029}} & \textbf{0.6001} & {\textbf{0.2450}}& {\textbf{0.0217}} & \textbf{{3.4956}} & \textbf{1.4889} &\textbf{1.4488} \\
\bottomrule
\end{tabular}
\vspace{5pt}
  \caption{Performance comparisons of \SeqFusion with a general model pre-trained on the entire pre-training data of model zoo. Specialized PTMs selected by \SeqFusion outperform the general model across all tasks. This highlights the importance of leveraging specialized models, as \SeqFusion dynamically selects PTMs tailored to the temporal patterns and forecasting requirements of the target time series, making full use of PTM resource and reducing extensive data collection.
}
  \label{tab:general}
  \vspace{-15pt}
\end{table*}

\noindent{\bf Model Zoo Family}. 
We conduct experiments to investigate how the composition and variety of the model zoo impact the performance of \SeqFusion. Firstly, we examine the effect of varying the number of PTMs in the model zoo. We increase the number of PTMs from the original 10 to 20, including 3 PTMs trained on the newly added Hospital~\cite{hospital} dataset and 7 PTMs trained on data sampled from the M3 and M4 datasets. We also explore the impact of different PTM architectures within the model zoo. Specifically, we mixed 10 PatchTST architecture (transformer-based) with 10 DLinear architecture (MLP-based). The final results are shown in the Table~\ref{tab:zoo_size}. 

\begin{table}[h]
  \centering
  \begin{tabular}{lcccc}
    \toprule
    \multirow{2}{*}{\textbf{Model Zoo}} & \multicolumn{4}{c}{\textbf{Datasets}} \\
    & ECL & ETTh1& Illness & Weather \\
    \midrule
    PatchTST x 10 & \textbf{0.6087} & \textbf{0.6001} &  1.4488 &3.6602\\
     + DLinear x 10  & 0.6156 & \textbf{0.6001 } & 1.4488 &5.1615\\ %& 0.6156 & 0.6297  & 1.4619 &5.1615\\ 
    PatchTST x 20 & 0.6281 & 0.6395  & \textbf{1.4471}&\textbf{3.3798}  \\ 
    \bottomrule
  \end{tabular}
  % \vspace{-2mm}
  \vspace{5pt}  
  % \caption{MSE performance of different model zoo family.\looseness=-1}
  \caption{MSE comparisons for different model zoo configurations. Increasing PTMs from 10 to 20, with models from Hospital, M3, and M4 datasets, improves performance on some datasets. Mixing architectures (PatchTST + DLinear) reduces accuracy.}
  \label{tab:zoo_size}
  \vspace{-15pt}
\end{table}

Increasing the number of PTMs from 10 to 20 shows a marginal improvement in performance on certain datasets, particularly the Illness dataset, where the MSE is slightly reduced from 1.4488 to 1.4471. This improvement can be attributed to the inclusion of PTMs trained on the Hospital dataset, which shares similar temporal and domain-specific characteristics with the Illness dataset. The knowledge transfer from related datasets (in this case, medical datasets with similar recording frequencies) appears to enhance SeqFusion’s ability to forecast accurately within this domain. In contrast, adding architectural diversity by mixing PatchTST (transformer-based) and DLinear (MLP-based) models leads to a decrease in performance, especially noticeable on datasets like Weather, where the MSE increases from 3.6602 to 5.1615. This suggests that while diversity in model architecture might introduce new perspectives, it can also disrupt the coherence of the model zoo, potentially introducing conflicting patterns or biases that reduce the ensemble’s effectiveness. Transformer-based architectures like PatchTST appear more suited to capturing temporal dependencies in these time-series datasets compared to MLP-based models like DLinear. Overall, these results indicate that to maximize \SeqFusion’s performance, it is essential to expand the model zoo with PTMs that capture diverse temporal patterns from relevant domains, rather than merely increasing architectural variety. In other words, incorporating PTMs trained on domain-similar datasets (e.g., Hospital for Illness) is more beneficial than mixing fundamentally different architectures, which may not align well with the specific temporal characteristics of the target datasets.

\begin{table*}[ht]
\centering
\setlength{\tabcolsep}{1.25mm} % Adjust the column separation
  
\begin{tabular}{lccccccc}
\toprule
\multirow{2}{*}{\textbf{Methods}} &  \multicolumn{7}{c}{\textbf{Downstream Target Dataset}}   \\
&
  \multicolumn{1}{l}{ECL} &
  \multicolumn{1}{l}{ETTh1} &
  \multicolumn{1}{l}{ETTh2} &
  \multicolumn{1}{l}{Exchange} &
  \multicolumn{1}{l}{illness} &
  \multicolumn{1}{l}{traffic} &
  \multicolumn{1}{l}{weather} \\
\midrule
TS2Vec~\cite{YueWDYHTX22}   & 0.6061 & \textbf{0.5723} & 0.2673  & 0.0247   & \textbf{3.3150} & \textbf{1.3748} & 1.4675\\
SimMTM~\cite{simmtm} w/o Trans.       & 0.6056  & 0.6164 & 0.2490  & 0.0239   & 3.4983 & {1.4360} & {1.4548} \\
SimMTM~\cite{simmtm} w/ Trans.         & {\textbf{0.6029}} & {0.6001} & {\textbf{0.2450}}& {\textbf{0.0217}} & {3.4956} & 1.4889 &\textbf{1.4488} \\
\bottomrule
\end{tabular}
\vspace{5pt}
  \caption{Performance comparisons of \SeqFusion using general extractor models with TS2Vec and SimMTM architectures (with/without transferability loss). While TS2Vec achieves comparable performance to SimMTM, the addition of transferability loss enhances performance across datasets.}
  \label{tab:ts2vec}
  \vspace{-15pt}
\end{table*}

\noindent{\bf How does transferability loss help?} To illustrate the effect of transferability loss in the training of the general extractor, we first compare the performance of \SeqFusion using general extractor models with/without transferability loss. In Table~\ref{tab:ts2vec}, it shows that the general extractor with transferability loss outperforms the one without consistently. In fact, the transferability loss helps PTMs trained from related datasets to be similar in the representation space. We use PCA to visualize all PTMs' representations extract from the general extractor model trained with the transferability loss. In Figure~\ref{fig:repr}, we can see that PTMs trained from the same source dataset are closer after using the transferability loss and vice versa. We also consider replace the architecture of the general extractor model with another time-series representation method TS2Vec~\cite{YueWDYHTX22}. In Table~\ref{tab:ts2vec}, we observe that TS2Vec demonstrates comparable performance to SimMTM. However, SimMTM with transferability loss shows a slight advantage on some more challenging tasks, such as ECL and weather dataset. This performance boost suggests that transferability loss helps align representations from related datasets, enhancing the model's ability to generalize across domains, which appears critical for tasks requiring cross-domain generalization.

% \begin{table}[t]
% \centering
  
% \begin{tabular}{lcccc}
% \toprule
% \multirow{2}{*}{\textbf{Methods}} &  \multicolumn{4}{c}{\textbf{Downstream Target Dataset}}   \\
% &
%   \multicolumn{1}{l}{ECL} &
%   \multicolumn{1}{l}{ETTh1} &
%   \multicolumn{1}{l}{Illness} &
%   \multicolumn{1}{l}{Weather} \\
% \midrule
% w/o Trans.        & 0.6056  & \textbf{0.5964 }     & 3.4983 & {1.4548} \\
% w/ Trans.             & {\textbf{0.6029}} & 0.6001& \textbf{{3.4956}}  &\textbf{1.4488} \\
% \bottomrule
% \end{tabular}
% % \vspace{-2mm}
% % \caption{MSE performance with or without transferability loss.}
% \vspace{5pt} 
% \caption{MSE compare with or without transferability loss.}
%   \label{tab:gt}
% \end{table}

\begin{figure}[ht]
\centering
\includegraphics[width=0.9\columnwidth]{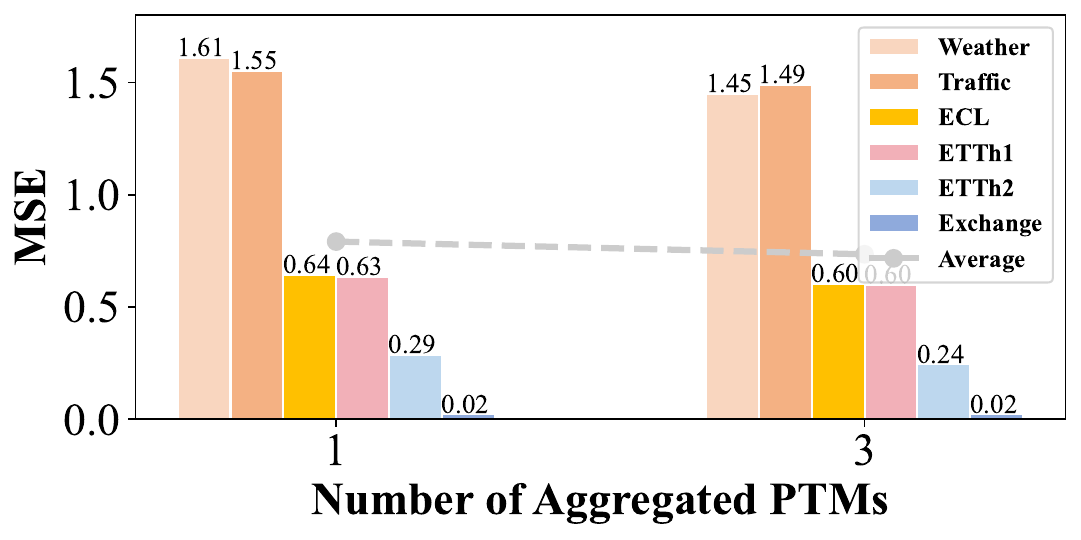} 
\caption{Performance of \SeqFusion with varying numbers of aggregated PTMs. Aggregating predictions from multiple PTMs reduces MSE across datasets.\looseness=-1}
\label{fig:ensemble}
\end{figure}

\noindent{\bf Enhancing forecasting with PTM Aggregation.} \SeqFusion is able to incorporate top suitable PTMs to improve forecasting. Figure~\ref{fig:ensemble} demonstrates that \SeqFusion benefits from aggregating predictions from multiple PTMs, with MSE decreasing as the number of aggregated PTMs increases. By combining predictions from multiple PTMs, SeqFusion effectively reduces the impact of any single model’s potential biases or errors, leading to more reliable final predictions. This ensemble approach allows SeqFusion to leverage the strengths of individual PTMs while mitigating weaknesses, resulting in an overall improvement in performance. The declining trend in average MSE as the number of PTMs increases suggests that aggregating predictions is a robust strategy for enhancing predictive accuracy across varied time-series forecasting tasks.

\section{Conclusion}
\label{Section7}

% \SeqFusion introduces a novel framework for zero-shot time-series forecasting by leveraging a diverse collection of pre-trained models (PTMs). By dynamically selecting suitable PTMs based on the target time series, \SeqFusion overcomes the limitations of traditional methods that require extensive training data. This approach enhances privacy, reduces storage costs, and increases model reuse flexibility. Through recursive forecasting and aggregated predictions, \SeqFusion achieves competitive accuracy across various datasets, demonstrating robustness and adaptability in data-limited environments. Experiments show \SeqFusion's potential to outperform state-of-the-art methods, making it a valuable contribution to time-series forecasting. Future research could optimize model selection and fusion strategies, integrate more PTMs, and expand applications to real-time forecasting and other domains.

\SeqFusion introduces a novel framework for zero-shot time-series forecasting that leverages the power of a diverse collection of pre-trained models (PTMs). Unlike traditional methods that require extensive in-task training data or generalized pre-trained models dependent on large-scale datasets, \SeqFusion dynamically selects and aggregates predictions from PTMs specifically tailored to the characteristics of the target time series. This approach enhances privacy, reduces storage costs, and increases model reuse flexibility. 
Through recursive forecasting and aggregated predictions, \SeqFusion achieves competitive accuracy across various datasets, demonstrating robustness and adaptability in data-limited environments. Experiments demonstrate \SeqFusion's potential to outperform state-of-the-art methods, showcasing its flexibility and practicality in diverse domains such as healthcare, finance, and environmental science.
A key innovation in \SeqFusion is its ability to extract these representations using only a small amount of data that does not compromise privacy. For example, in financial applications like stock market forecasting, where proprietary trading data must remain confidential, \SeqFusion can rely on limited, non-sensitive indicators such as sector-level averages or public historical stock prices to extract meaningful representations. These minimal yet informative datasets allow the feature extractor to align target time series with suitable PTMs without requiring access to detailed or proprietary raw data.

The success of \SeqFusion highlights the potential for PTM-based approaches in time-series forecasting. Future work could explore more sophisticated model selection and fusion strategies, such as adaptive weighting or task-specific optimization. Additionally, integrating more PTMs from diverse domains and the extension of \SeqFusion to handle real-time forecasting tasks could further enhance its applicability. By demonstrating the feasibility of sharing and reusing PTMs for time-series applications, \SeqFusion paves the way for advancing zero-shot forecasting methodologies and their adoption in industry and research.

\bibliographystyle{IEEEtran}
\bibliography{main.bib}

\end{document}